# Augmented Skeleton Space Transfer for Depth-based Hand Pose Estimation

Seungryul Baek
Imperial College London
s.baek15@imperial.ac.uk

Kwang In Kim
University of Bath
k.kim@bath.ac.uk

Tae-Kyun Kim
Imperial College London
tk.kim@imperial.ac.uk

## Abstract

*Crucial to the success of training a depth-based 3D hand pose estimator (HPE) is the availability of comprehensive datasets covering diverse camera perspectives, shapes, and pose variations. However, collecting such annotated datasets is challenging. We propose to* complete *existing databases by generating new database entries. The key idea is to synthesize data in the skeleton space (instead of doing so in the depth-map space) which enables an easy and intuitive way of manipulating data entries. Since the skeleton entries generated in this way do not have the corresponding depth map entries, we exploit them by training a separate hand pose generator (HPG) which synthesizes the depth map from the skeleton entries. By training the HPG and HPE in a single unified optimization framework enforcing that 1) the HPE agrees with the paired depth and skeleton entries; and 2) the HPG-HPE combination satisfies the cyclic consistency (both the input and the output of HPG-HPE are skeletons) observed via the newly generated unpaired skeletons, our algorithm constructs a HPE which is robust to variations that go beyond the coverage of the existing database.*

*Our training algorithm adopts the generative adversarial networks (GAN) training process. As a by-product, we obtain a hand pose discriminator (HPD) that is capable of picking out realistic hand poses. Our algorithm exploits this capability to refine the initial skeleton estimates in testing, further improving the accuracy. We test our algorithm on four challenging benchmark datasets (ICVL, MSRA, NYU and Big Hand 2.2M datasets) and demonstrate that our approach outperforms or is on par with state-of-the-art methods quantitatively and qualitatively.*

## 1. Introduction

Estimating the 3D pose of a hand from a single depth map finds numerous applications in human-computer interaction, computer graphics, and virtual & augmented reality and it has emerged as a key problem in computer vision [44, 32, 51, 59, 11, 53, 30, 4, 13, 57, 37, 52, 26, 60, 48, 46, 10, 33, 19, 41, 20, 14, 18, 22, 9, 58].

Hand pose estimation is a challenging problem. Typical application scenarios of hand pose estimation require identifying almost the same number of parameters ($\approx 100$; see Fig. 2) as human body pose. However, in contrast with body pose estimation where subjects are typically isolated and in the upright position, hands exhibit frequent and severe self-occlusions, and furthermore, it is not straightforward to define *canonical views* as hands are captured in a wide range of equally likely camera perspectives. Therefore, the ability to operate reliably under varying camera perspectives, poses, and shapes (subject identities) is crucial to successful hand pose estimation.

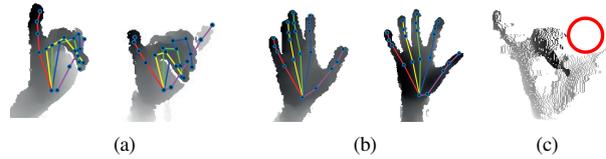

Figure 1: Similar hand skeletons overlaid with the corresponding depth maps in (a) camera perspective and (b) shape (subject) variations: Note that slight variations in hand skeletons are instantiated to significantly different depth maps. It is much easier to edit hands in the skeleton spaces as directly manipulating depth values is challenging. Further, as the original depth map can exhibit self-occlusions, naïve 3D rotations introduce holes for large rotation angles (c): The occluded (empty) thumb region is highlighted with a red circle (60-degree rotation from (a) left).

A straightforward approach to construct such a robust hand pose estimator (HPE) might be to train it on a large dataset that covers such variations. However, as far as we are aware, existing datasets are limited in the coverage of camera viewpoint, shape, and/or pose variations (see Sec. 3).

When the data space is visualized using the ground-truth annotations of hand poses, shapes, and camera perspectives in such databases, one can identify the *missing* regions in the space, *e.g.* camera perspectives that are not covered by the database (see Fig. 4). This motivates a study to complete the dataset by synthesizing new skeleton and depth map pairs. Using synthetic data as one such approach, typically requires physical or statistical hand models. Furthermore, even with advances in graphics, synthetic depth maps exhibit observable differences from real data [31]. An alternative is *simple* depth map and skeleton pair manipulation, *e.g.* by in-plane rotations and translations. As shown in our experiments, while this way of augmenting data helps, the resulting database coverage, however, is limited. Extending this data manipulation approach to non-trivial variations in shape and 3D camera view is challenging: Directly changing the depth values can easily generate unrealistic hand shapes as data entries in depth maps are highly structured and correlated. Furthermore, similar hand skeletons can be instantiated as significantly diffident depth maps indicating the inefficiency of directly manipulating them (see Fig. 1).

In this paper, we present an algorithm that mitigates these limitations by augmenting the camera views and shapes. In training, we synthesize unseen human hands in the *skeleton space* and *transfer* them to synthetic depth maps: This helps to avoid the challenge in manipulating the depth maps (which vary drastically with respect to mild variations in viewpoints and shapes; see Fig. 1) and provide an easy and intuitive way to close the gaps in the data space by editing existing data points. To facilitate the transfer of generated skeletons to depth maps, we introduce two new data processing networks: Inspired from the recent success of 2D/3D image generation [23, 35, 34, 45, 55, 56, 17, 7], we train the hand pose generator (HPG) that transfers input skeletons to corresponding depth maps. As in generative adversarial networks (GANs) [12, 27], we train the second, hand pose discriminator (HPD) that distinguishes real depth maps from these synthesized by the HPG. Combining and jointly training HPG, HPE, and HPD enable the automatic transfer of the augmented skeletons to depth maps by enforcing the consistency over existing paired skeleton and depth map database entries and the self-consistency over unpaired augmented skeletons. The HPD's ability (combined with HPG) to pick out realistic human hands can also be used in testing: During testing, we synthesize multiple hand pose hypotheses out of the initial HPE prediction, and generate the final refined prediction by combining them using the HPG and HPD as a prior.

To summarize, we contribute by 1) a new hand pose generator (HPG) and estimator (HPE) combination that enables to exploit both existing paired skeletons and depth map entries and newly synthesized depth maps in a single unified framework; 2) a strategy that refines the HPE prediction during testing by generating multiple pose hypotheses and combining them using HPD and HPG as a prior. In the experiments with four challenging datasets, we demonstrate that our robust algorithm outperforms or is on par with state-of-the-art algorithms on each dataset.

## 2. Related work

**Hand pose generator (HPG)-guided approaches.** The recent success of image generation networks has demonstrated the use of generative networks to guide the training of estimators. Oberweger *et al.* [21] used HPG to synthesize depth maps from estimated skeletons. This depth map is then compared with the original, input depth map to quantify the difference and iteratively refine the skeleton estimate. Wan *et al.* [51] proposed a semi-supervised learning framework that uses HPG to exploit the unpaired depth maps and learns a latent space shared by hand poses and depth maps. Our approach is inspired from these approaches but differs in details as they are aligned with a different motivation: The primary goal of our algorithm is to enrich existing datasets by augmenting them. Since augmenting data in skeleton space is much easier, our algorithm focuses on the capability of transferring the augmented skeletons to depth maps. In contrast, existing algorithm focus on exploiting data entries within their limits, *e.g.* Wan *et al.*'s exploit unpaired depth maps. Also, our algorithm is complementary to Oberweger *et al.*'s algorithm as it uses the HPG only in the testing phase while ours jointly train HPE and HPG, and uses them in both training and testing.

Recent progresses in graphics have made the use of synthetic data a competitive alternative to building expensive annotations. Still there is an observable gap between real and synthetic data entries. Shrivastava *et al.*'s algorithm focuses on reducing such gap [31]: Their conditional HPG receives a depth map rendered from a physical hand model, and generates a more realistic one simulating non-smooth observations at object boundaries. Combined with a generative hand model, this algorithm can be used to fill in the missing regions in the datasets similarly to ours. Our algorithm makes a complementary approach that does not require a physical or statistical hand model.

**Multi-view/shape approaches.** An alternative to constructing robust HPE under view-point and shape variations is to apply multi-view approaches. A depth map can be regarded as a projection of a 3D object onto a view plane, partially losing 3D structural information. In this respect, exploiting additional views of an underlying 3D object have shown to improve the performance in related applications (*e.g.* human action recognition [29, 28, 2, 3] and 3D object recognition [50, 25, 54, 38]). Applying to hand pose estimation, Simon *et al.*'s multi-view boostrapping algorithm adopts a multi-camera system in training where each view-dependent initial estimate is iteratively triangulated in 3D and refined [32]. While this approach has demonstrated the potential of multi-view approaches, it is non-trivial to apply to single depth map-based systems. Furthermore, in general multi-view approaches cover only viewpoint variations. Ge *et al.* proposed an algorithm that simulates the multi-view approach by generating multiple 2D views from a single depth image [10]: They first estimate a 3D point cloud from the input depth map, and then project them onto three 2D view planes ($x-y$, $x-z$, and $z-x$ panes). The 3D pose estimates are then constructed by applying 2D convolutional neural networks (CNNs) to each views followed by multiple view scene fusion. They extended this idea to generate multi-view 3D cliometric forms and fuse them via a single neural network aggregrator [11]. This type of approach helps to disambiguate between similar poses to a certain degree. However, their 3D reconstruction ability is inherently limited based on the initial point clouds estimated again from a single view-dependent depth map. The performance of these approaches, therefore, depend on the richness of the dataset.

The robustness over the shape variation is also important in hand pose estimation. While the estimation accuracy is affected by hand shapes, collecting comprehensive datasets to train a robust estimator is challenging. An alternative is to apply explicit 3D hand model-based approaches [5, 36, 40, 16, 46] that simultaneously optimize the shape, viewpoint, and pose parameters of the model. These algorithms, however, require solving a complex optimization problem during testing.

**Pose refinement.** During training, our algorithm constructs an auxiliary skeleton discriminator (HPD) that allows hypothesis testing of whether a given skeleton is plausible or not (as generated by the HPG). Combined with the HPG, HPD can thus act as a prior on the target hand pose space. This leads to a framework that *refines* the initial skeleton estimated by the HPE as motivated by the success of existing refinement-based approaches [39, 21, 53]. Sun *et al*. [39] propose a linear sequence (cascade) of weak regressors that are trained to output residuals to guide input finger joints towards their ground-truths. They iteratively refine the once estimated skeleton joints by updating their palm joints (global pose) and the finger joints thereafter (local pose). Oberweger *et al*.'s approach also iteratively refines the initial hand pose estimates as guided by the HPG [21]. Wu *et al*.'s algorithm constructs a skeletal Gaussian mixture model which acts as a prior. The initial pose estimates are then refined by combining the prior with 2D re-projection and temporal consistency likelihood [53].

Compared to prior works, our refinement method differs in that it exploits the information on 1) the augmented skeletons and the corresponding transferred depth maps via feedback from the trained discriminator HPD$_Y$ and 2) multiple viewpoints hypothesized by manipulating the initial estimates.

## 3. Pose estimation by skeleton set augmentation

Given a database of input depth maps and the corresponding ground-truth hand pose annotations $P = \{(\mathbf{x}_i, \mathbf{y}_i)\}_{i=1}^{l} \subset X \times Y$, our goal is to construct a **hand pose estimator (HPE)** $f^E : X \to Y$ that recovers the underlying pose $\mathbf{y}'$ of an unseen test depth map $\mathbf{x}'$. When the *paired* dataset $P$ is large enough to cover variations in poses, shapes, views, *etc*., a straightforward approach to train such a HPE is to minimize the mean squared loss over $P$:

$$\mathcal{L}_E(f^E) = \sum_{i=1}^{l} \|f^E(\mathbf{x}_i) - \mathbf{y}_i\|_2^2. \quad (1)$$

For this baseline, we use the convolutional neural network (CNN) architecture in [59]: Each input depth map $\mathbf{x}$ is presented as a $96 \times 96$-dimensional array while for the output $\mathbf{y}$, we adopt the 63-dimensional skeletal pose vector representing the $(x,y,z)$-coordinate values of 21 hand joints (Fig. 2).

Unfortunately, existing datasets do not comprehensively cover the wide variety of hand shape and views. Therefore, we explicitly fill-in these missing regions by synthesizing data entries in the skeleton space $Y$. Once such *unpaired* skeletal poses $U = \{\mathbf{z}_i\}_{i=1}^{u}$ are synthesized, training the HPE is performed based on a combination of the standard estimation error $\mathcal{L}_E$ via $P$ (Eq. 4) and the cyclic consistency requirements induced from $U$ (see Fig. 5 and Eq. 7). To facilitate this process, we train a hand pose generator (HPG) $f^G : Y \to X$ that receives a skeleton (either $\mathbf{y}$ or $\mathbf{z}$) and synthesizes the corresponding depth map $\mathbf{x}$. Note our skeletal representation incorporates camera perspectives: 63-dimensional skeletal pose values are assigned based on the coordinate system defined by the views.

### 3.1. Skeleton set augmentation

**Skeletal hand shape model.** We use the 21 joint-based skeletal hand shape model proposed in [59] (Fig. 2). This model represents a human hand based on 25 joint angles and the lengths of 20 bones connecting joint pairs: Each finger pose is represented as 5 angles (twist angle, flexion angle, abduction angle for the MCP joint and flexion angles for the DIP and PIP joints) and 4 bone lengths.[1]

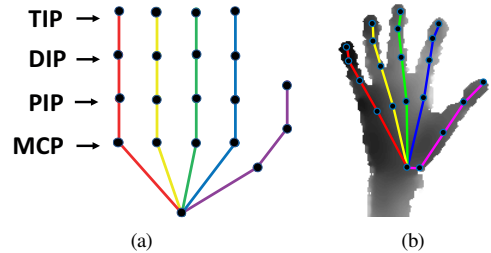

Figure 2: Our skeletal hand model (a) consists of 21 joints [59]: one for wrist and four for each finger. Each finger has five degrees of freedom: flexion for DIP and PIP, flexion, abduction and twist for MCP. (b) a skeleton overlaid on the underlying depth map.

**Hand datasets: obtaining $P$.** The performance of HPE depends on its training datasets. The *Big Hand 2.2M* dataset collected by Yuan *et al*. [59] is the largest dataset including 2.2 million frames extracted from sequences of $\binom{32}{2} = 496$ transitions between $2^5 = 32$ extreme hand poses. While it provides a comprehensive hand pose coverage, *Big Hand 2.2M* still lacks the variety in hand shapes (only 10 subjects) and in camera views (see Fig. 3). Other popular datasets include *ICVL* [42], *NYU* [49] and *MSRA* [39]. The *ICVL* benchmark [42] includes only 1 subject and provides a limited coverage of viewpoints and poses consisting of 17,604 frames. The *NYU* dataset provides a broader range of viewpoints (81,009 frames) but with limited shape variations (one subject for training and another subject for testing). The *MSRA* [39] benchmark is similar in scale to *NYU* (76,375 frames) but with a more comprehensive viewpoint coverage. However, its shape and pose variations are limited to 9 subjects and 17 discretized poses per subject, respectively.

**Skeleton augmentation: constructing $U$.** For each of the four datasets aforementioned, we enlarge its skeleton space coverage by adding variations in viewpoints and shapes. We do not consider pose (*i.e*. articulation) augmentation as we observed in preliminary experiments that synthesizing realistic hand poses without having access to statistical or physical hand models is challenging.

New camera perspectives (viewpoints) of an existing skeleton entry are synthesized by applying (3D) rotations along $y-z$ and $x-z$ panes, prescribed by the corresponding rotation degrees $\theta_1$ and $\theta_2$. In-plane rotations (along $x-y$ pane) can also be considered but we exclude them in the skeleton augmentation process as the corresponding paired data is straightforwardly constructed by rotating the skeleton and depth map pairs. In the experiments (Table 1(b)) we demonstrate that both simple in-plane rotations and our skeletal augmentation help improve the performance and furthermore, they are complementary: The combination of the two data augmentation modes is better than either taken alone.

---

[1]TIP stands for the finger tip. MCP, PIP, and DIP represent the MetaCarpoPhalangeal, Proximal InterPhalangeal, and Distal InterPhalangeal joints, respectively.

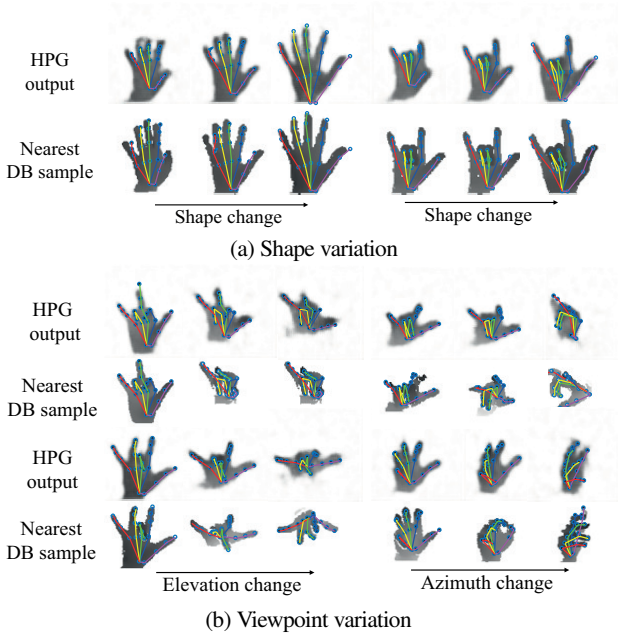

(a) Shape variation

(b) Viewpoint variation

Figure 3: Synthesized skeletons $\mathbf{y}'$ overlaid on the depth maps $\mathbf{z}'$ transferred by HPG ($\mathbf{x}' = f^G(\mathbf{y}')$). These new data entries *augment* the coverage of the database: The nearest skeletons (and the paired depth maps) in the database deviate significantly from the query synthesized skeletons.

Adopting existing models from human hand shape analysis [8, 6], we characterize hand shapes based on the width to length ratio of each finger. Accordingly, new skeletons are generated by varying the finger lengths of existing data entries as measured in Euclidean distances between TIP to DIP, DIP to PIP and PIP to MCP while fixing the palm (see Fig. 2). While fixing 6 palm positions, we first identify 5 angles (*i.e.* flexion angles for TIP, DIP and PIP and twist angle, flexion angle abduction angle for the MCP) and 3 bone lengths (distances from MCP to PIP, from PIP to DIP and from DIP to TIP) for each finger and given them, reconstruct each finger by rotating/translating their end points to attach them to the palm. We assume that the ratios of finger lengths are fixed and thus only the bone length of each finger is manipulated by multiplying a global constant $\tau$ in the above reconstruction process.

The variation parameters $\theta_1$, $\theta_2$, and $\tau$ are sampled from Gaussian distributions: $\mathcal{N}(1.0, 0.5^2)$ for $\tau$ and $\mathcal{N}(0, \frac{\pi}{4}^2)$ for $\theta_1$ and $\theta_2$. While in general, more sophisticated data manipulation strategies can be adopted, our preliminary visual evaluation on a small sample revealed that skeletons generated in this way look realistic. Figure 3 shows example skeletons (overlaid on the corresponding depth maps synthesized by HPG; see Sec. 3.2) generated from this process. Note that the alternative way of directly manipulating depth maps could be much more challenging as the variables are highly structured and correlated, and therefore naïvely manipulating depth pixels would lead to unrealistic hand shapes. We apply the manipulation process $M$ times for each database entry constructing an unpaired database $U$ ($|U| = M|P|$). Table 1(b) shows

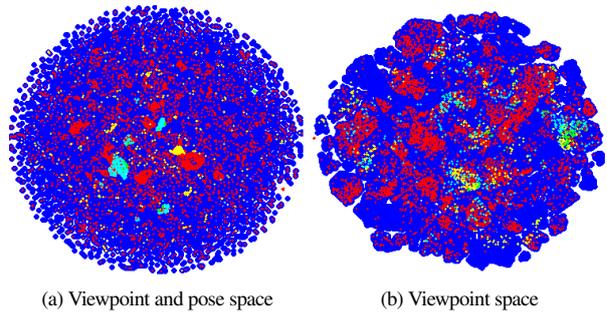

(a) Viewpoint and pose space     (b) Viewpoint space

Figure 4: t-SNE embeddings of skeletal poses of *Big Hand 2.2M*, *ICVL*, *NYU*, *MSRA* datasets, and our augmented skeletons. Each dataset covers up to a certain degree of shape and viewpoint variations but none of them is *comprehensive* as indicated by the presence of empty space between different clusters. Our data augmentation process fills in the space and provides a more comprehensive coverage of viewpoints and poses. To experience the full detail of this figure, readers are advised to view the electronic version.

the effect of varying $M$ on the final pose estimation performance.

Figure 4 visualizes the results of skeleton augmentation: We observe that even the biggest *Big Hand 2.2M* dataset is far from being fully covering the wide variations in shapes and camera viewpoints as evidenced by almost 10-times larger area coverage accomplished by our augmented dataset.

### 3.2. Transferring skeletons to depth maps

**Hand pose generator (HPG).** Our HPG $f^G$ synthesizes a depth map $\mathbf{x}$ given the input skeleton parameters $\mathbf{y}$. We adopt Pathak *et al.*'s conditional GAN architecture [24] that combines both $L_2$-loss and adversarial loss (Eq. 3): The $L_2$ loss (defined via $P$) measures the deviation of the synthesized depth maps from the ground-truths while the adversarial loss generates data distribution and helps the generator to synthesize more plausible data samples.

**Hand pose discriminator (HPD).** We construct auxiliary models that provide feedback on the quality of synthesized data. To leverage the cyclic nature of HPE and HPG combinations (*i.e.* $f^E(f^G)$ and $f^G(f^E)$ map $Y$ to itself and $X$ to itself, respectively; see the next paragraph), we train two such discriminators: The depth hand pose discriminator (HPD$_X$) $f^{D_X}$ is the same as the standard GAN discriminator; It outputs 1 for real data entries and 0 for the synthesized entries. The role of skeleton hand pose discriminator (HPD$_Y$) $f^{D_Y}$ is to decide whether the estimated finger joints conform the human skeleton model (Fig. 2). It aims to accept original skeletal entries $\mathbf{y}$ in $P$ as well as the augmented entries $\mathbf{z}$ in $U$, while rejecting outputs from the HPE. Therefore, incorporating $f^{D_Y}$ into the joint GAN training enables us to steer the training of the generator $f^G$ towards the skeletal poses that were not covered in the original dataset $P$.

**Training HPG and HPE.** Our goal is to jointly train HPG and HPE by fully exploiting the paired data $P$ as well as the augmented unpaired data $U$. Since skeletons $\mathbf{z}$ in unpaired data $U$ have no corresponding depth maps $\mathbf{x}$ for explicit supervision, our training algorithm adopts cyclic consistency: When the depth map $f^G(\mathbf{y}) \in X$ generated from a hypothesized input pose $\mathbf{z}$ is fed to the HPE, the resultant $f^E(f^G(\mathbf{z})) \in Y$ should be *similar* to $\mathbf{z}$. Similarly, the HPE result $f^E(\mathbf{x}) \in Y$ for an input depth map $\mathbf{x}$ can be fed to HPG and the resulting simulated depth map $f^G(f^E(\mathbf{x})) \in X$ should be similar to the original input $\mathbf{x}$.

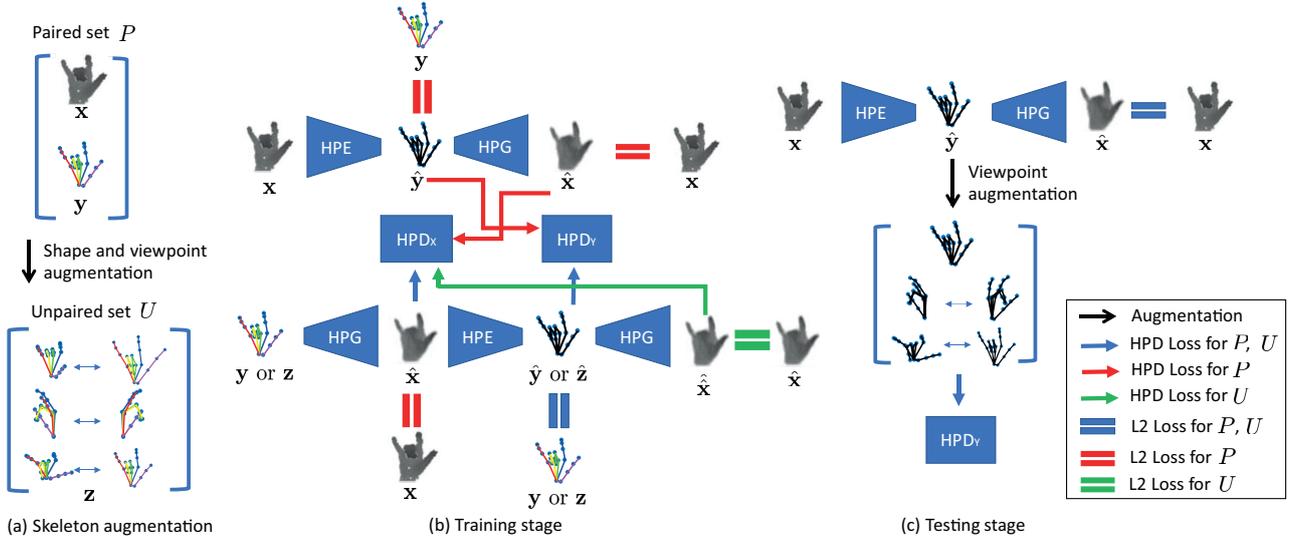

Figure 5: Schematic diagrams of our algorithm. (a) Manipulating skeletons is easier than manipulating depth maps; (b) During training, HPE, HPG, HPD$_X$, and HPD$_Y$ are optimized by 1) reducing the classical training error of HPE induced via $P$; 2) enforcing the cyclic consistency of HPE-HPG combination $f^E(f^G):Y \to Y$, HPG-HPE combination $f^E(f^G):X \to X$ on $P$ as well as the HPG-HPE-HPG consistency on unpaired data $U$; (c) In testing, our algorithm refines the initial hand pose prediction as guided by HPG and HPD$_Y$ as a prior. In the diagram, Red and Green lines represent interactions with the paired set $P$ and unpaired set $U$, respectively. The Blue lines represent interactions with both $U$ and $P$.

Accounting for these requirements, our energy $\mathcal{L}$ consists of four components: The individual losses for HPE and HPG, respectively, and the consistency losses for $f^E(f^G)$ and $f^G(f^E)$ measured based on the paired ($P$) and unpaired ($U$) data:

$$\mathcal{L}(f^G, f^E, f^{D_X}, f^{D_Y}) = \mathcal{L}_G(f^G, f^{D_X}) \\ + \mathcal{L}_E(f^E, f^{D_Y}) + \lambda(\mathcal{L}_P(f^E, f^G) + \mathcal{L}_U(f^E, f^G)). \quad (2)$$

The individual losses $\mathcal{L}_G$ and $\mathcal{L}_E$ are defined as:

$$\mathcal{L}_G(f^G, f^{D_X}) = \mathbb{E}_\mathbf{x}[\log f^{D_X}(\mathbf{x})] \\ + \mathbb{E}_\mathbf{y}\left[\log(1 - f^{D_X}(f^G(\mathbf{y}))) + \|f^G(\mathbf{y}) - \mathbf{x}\|_2^2\right] \quad (3)$$

$$\mathcal{L}_E(f^E, f^{D_Y}) = \mathbb{E}_\mathbf{y}[\log f^{D_Y}(\mathbf{y})] \\ + \mathbb{E}_\mathbf{x}\left[\log(1 - f^{D_Y}(f^E(\mathbf{x}))) + \|f^E(\mathbf{x}) - \mathbf{y}\|_2^2\right] \quad (4)$$

with the expectations $\mathbb{E}_\mathbf{x}$ and $\mathbb{E}_\mathbf{y}$ taken respectively under the empirical marginal distributions $p_X(\mathbf{x})$ and $p_Y(\mathbf{y})$ (defined by $P$). The consistency losses for the paired ($\mathcal{L}_P$) and unpaired ($\mathcal{L}_U$) datasets are respectively given as:

$$\mathcal{L}_P(f^E, f^G) = \mathbb{E}_\mathbf{y}\left[\log f^{D_X}(\mathbf{x}) + \|f^E(f^G(\mathbf{y})) - \mathbf{y}\|_2^2\right] \\ + \mathbb{E}_\mathbf{x}\left[\|f^G(f^E(\mathbf{x})) - \mathbf{x}\|_2^2 + \log(1 - f^{D_X}(f^G(f^E(\mathbf{x}))))\right] \\ + \mathbb{E}_\mathbf{y}\left[\log f^{D_Y}(\mathbf{y}) + \log(1 - f^{D_Y}(f^E(f^G(\mathbf{y}))))\right], \quad (5)$$

$$\mathcal{L}_U(f^E, f^G) = \mathbb{E}_\mathbf{z}\left[\log f^{D_Y}(\mathbf{z}) + \log(1 - f^{D_Y}(f^E(f^G(\mathbf{z})))) \\ + \|f^E(f^G(\mathbf{z})) - \mathbf{z}\|_2^2 + \|f^G(f^E(f^G(\mathbf{z}))) - f^G(\mathbf{z})\|_2^2 \\ + \log(1 - f^{D_X}(f^G(\mathbf{z}))) + \log(1 - f^{D_X}(f^G(f^E(f^G(\mathbf{z})))))\right], \quad (6)$$

where the expectation $\mathbb{E}_\mathbf{z}$ is taken over the empirical distribution $P(\mathbf{z})$ of the augmented skeletons $\mathbf{z} \in Y$ (defined by $U$).

*Discussions.* The cyclic consistency of $f^E(f^G)$ and $f^G(f^E)$ combinations on unpaired data $U$ are inspired by the consistency loss of cyclic GAN [61] that exploits the mutual consistency of two transfer functions $f^E: X \to Y$ and $f^G: Y \to X$ (adapted to our problem setting). Indeed, the last terms in $\mathcal{L}_P$ and $\mathcal{L}_U$ are smooth versions of the consistency loss in [61]. However, the main goal of cyclic GAN training is to automatically infer the correspondences between two unpaired sets $U_X$ and $U_Y$ (adapted to our problem setting) without having to use explicitly paired data. Therefore, in [61], the consistency is enforced on two sets of unpaired data $U_X$ and $U_Y$. Our algorithm aims to achieve a similar goal but it is provided with a small paired dataset $P$ plus a large unpaired dataset $U := U_X$ only in the skeleton space (as augmenting data in $Y$ is challenging). Therefore, our algorithm indirectly induces the consistency of $f^E(f^G)$ and $f^G(f^E)$ by putting a consistency loss over a complete circle (the third and

**Algorithm 1:** Training process for HPG and HPE

**Input**: Depth map and skeleton pairs $P=\{(\mathbf{x}_i,\mathbf{y}_i)\}_{i=1}^{l}$ and unpaired skeletons $U=\{\mathbf{z}_i\}_{i=1}^{u}$; Hyper-parameters: the number $T$ of epochs and the size $N'$ of mini-batch;

**Output**:
HPE $f^E$, HPG $f^G$, HPD$_X$ $f^{D_X}$, and HPD$_Y$ $f^{D_Y}$.

**Initialization**:
randomly allocate parameters of $f^E$, $f^G$, $f^{D_X}$, and $f^{D_Y}$;

**for** $t=1$ **to** $T$ **do**
    **for** $n=1$ **to** $N'$ **do**
        Evaluate (feed-forward) $f^G$ and $f^{D_X}$ and their respective gradients $\nabla f^G$ and $\nabla f^{D_X}$ on $P$ (Eq. 3);
        Evaluate $f^E$, $f^{D_Y}$, $\nabla f^E$, and $\nabla f^{D_Y}$ (Eq. 4);
        Evaluate $f^E(f^G)$, $f^G(f^E)$, $f^{D_Y}$, and their gradients on $U$ and $P$ (Eqs. 5 and 6);
        Update $f^G$, $f^{D_X}$, $f^{D_Y}$ combining the calculated gradients.
        Evaluate $f^E(f^G)$, $f^G(f^E)$, and $f^{D_Y}$, and their gradients on $U$ and $P$ (Eqs. 5 and 6) and update $f^E$ accordingly.
    **end**
**end**

last terms in the $\mathcal{L}_U$ expectation: Eq. 6):

$$\mathbf{z}\xrightarrow{f^G}\hat{\mathbf{x}}\xrightarrow{f^E}\hat{\mathbf{z}}\xrightarrow{f^G}\hat{\hat{\mathbf{x}}}\approx\hat{\mathbf{x}}. \quad (7)$$

Furthermore, our consistency losses incorporate the contributions from the discriminators $f^{D_X}$ and $f^{D_Y}$. This helps in decoupling the updates of $f^E$ and $f^G$ (within each mini-batch) : We empirically observed that simultaneously updating $f^G$ and $f^E$ by combining all gradients is prone to overfitting, *i.e.* the resulting HPE-HPG combination memorizes the depth map entries in $U$ but it cannot faithfully re-generate over unseen depth maps. This can be attributed to the significantly larger dimensionalities of the depth map space $X$ (*e.g.*, $96^2$) than the parameterized skeleton space $Y$ (63): By simultaneously updating $f^G$ and $f^E$, the algorithm tends to emphasis the losses observed at $X$. These different scaling behaviors of $X$- and $Y$-losses can be addressed by explicitly scaling them, but it requires tuning a separate scaling parameter. Instead, our algorithm avoids such degeneracy by simply decoupling the updates of $f^E$ and $f^G$. Algorithm 1 and Fig. 5(b) summarizes the training process.

Our model is trained using the Adam optimizer [15] with its default parameters: $\beta_1 = 0.9$, $\beta_2 = 0.999$, and $\epsilon = 10^{-8}$. The learning rate and the regularization parameter $\lambda$ (Eq. 2) are fixed at $10^{-4}$ and $10^{-4}$, respectively based on cross-validation on *Big Hand 2.2M* dataset, which are fixed throughout the entire experiments (over other datasets).

**Refining predictions at testing.** Once the hand pose estimator $f^E$ is trained, it can be directly applied to an unseen input depth map $\mathbf{x}'$ to generate the output pose estimate $\mathbf{y}'$. However, during the training of HPE-HPG combination, we constructed an auxiliary hand pose discriminator (HPD) $f^{D_Y}$ that (combined with the HPG) can identify realistic skeleton configurations. Therefore, we refine the initial estimate $\mathbf{y}'$ as guided by HPG and HPD: Our initial result $\mathbf{y}'$ is updated using the gradient back-propagated from the HPG and HPD$_Y$ (see Fig 5(c)):

$$\mathbf{y}^*=\mathbf{y}'-\gamma\nabla\big(-f^{D_Y}(\mathbf{y}')+\lambda_{ref}\|f^G(\mathbf{y}')-\mathbf{x}'\|_2^2\big). \quad (8)$$

where $\gamma = 10^{-5}$ and $\lambda_{ref} = 0.01$ are fixed for all datasets by cross-validation on *Big Hand 2.2M*. This corresponds to a (computationally cheap) single step of energy minimization. In this way, the refined skeleton joints move towards matching the distribution of plausible skeleton joints.

*Multi-view gradient ensemble.* Throughout the training process, $f^{D_X}$ and $f^G$ have access to the augmented skeletons $U$ and the corresponding transferred depth maps $f^G|_U$, covering a variety of viewpoints. We generalize our refinement strategy to a multi-view scenario by exploiting this accumulated *multi-view knowledge*: First, our refinement step generates $R$-different views of the initial estimate $\mathbf{y}'$ by rotating it $R$-times, similarly to the skeleton augmentation process in training. These multiple view hypotheses are then fed to $f^{D_Y}$ and $F^G$ to generate the respective gradient updates (Eq. 8). The final prediction is then obtained by rotating back the updated results $\{\mathbf{y}^*\}$ to the original views and taking the average. For the rotated skeletons, $\lambda_{ref}$ is set to 0 in Eq. 8 as they do not have the corresponding rotated depth maps. We fix $R$ at 50 trading off the run-time complexity with the accuracy.

## 4. Experiments

We evaluate our algorithm on four depth-based hand pose estimation datasets: *Big Hand 2.2M*, *MSRA*, *ICVL*, and *NYU*. Each dataset differs in their intended use cases and properties, and therefore, we adopt different experimental settings per dataset.

**Setup.** For *Big Hand 2.2M*, we use the experimental settings proposed by the authors of the dataset, Yuan *et al.* [59] (see Sec. 3.1 for the discussion on this dataset): We use 90% and 10% of database entries for training and validation, respectively. For testing, 37,000 frames captured from a subject not contained in the training/validation sets are used. For the *MSRA* dataset containing 9 subjects, we follow the setting of [39, 10, 11] where the depth map and skeleton pairs of 8 subjects are used for training and the remaining data pairs (of one subject) are used for testing. We repeated the experiments for each subject left out and the observed accuracies are averaged. For both *Big Hand 2.2M* and *MSRA*, the output space represents 21 skeleton joints. For the *ICVL* dataset, we use the ground-truth annotations (21 skeleton joints) provided by Tang *et al.* [43, 59]. This model differs from the 16-joint model provided by the authors of the original data [42]. While in principle, our model can be applied to any output configurations, we adopt this 21-joint model for simplicity of evaluation and to facilitate the cross-dataset transfer experiments (to be discussed shortly). The combination of training and testing sets also follows from [43, 59]: 16,008 frames are used for training while the remaining 1,596 frames are used in testing. For the *NYU* dataset, we adopt the experimental settings of [21, 31, 51]: 72,757 frames and 8,252 frames are used for training and testing, respectively. Following their experimental settings, we estimate the 14 target skeleton joints (out of 36 joints in the original datasets). For comparison, we adopt several state-of-the-art methods that share the same evaluation protocol: For *Big Hand 2.2M*, we compare with CNN estimator employed by Yuan *et al.* [59] which constitutes our baseline HPE. We also evaluate two existing generative hand model-based approaches: FORTH [22] and Intel RealSense SR300 camera [1]. On *ICVL*, we compare with Sun *et al.*'s cascaded refinement algorithm (denoted as Sun *et al.*) [39] and Tang *et al.*'s hierarchical decision forests-based algorithm (Tang *et al.*) [43] which constitutes the state-of-the-art on this benchmark. For *MSRA*, we compare with two state-of-the-art methods, Ge *et al.* [10] and Ge *et al.* [11]). Both algorithms [10, 11] adopt the multi-view approach and therefore, they are especially effective for *MSRA* that covers diverse view points. For the *NYU* dataset, in addition to Sun *et al.*'s cascade algorithm [21] and Ge *et al.*'s multi-view approach [51], we compare with two generative model-based algorithms (Wan *et al.* [51]) constituting the-state-of-the-art on this dataset. All components of our networks were implemented with the Torch library and they are trained and evaluated on an Intel 3.40 GHz i7 machine with two NVIDIA GTX 1070 GPUs. Training our network on 10 times augmented *Big Hand 2.2M* dataset takes 3-4 days (100 epochs). At testing, our algorithm processes 300 frames per second using the GPU.

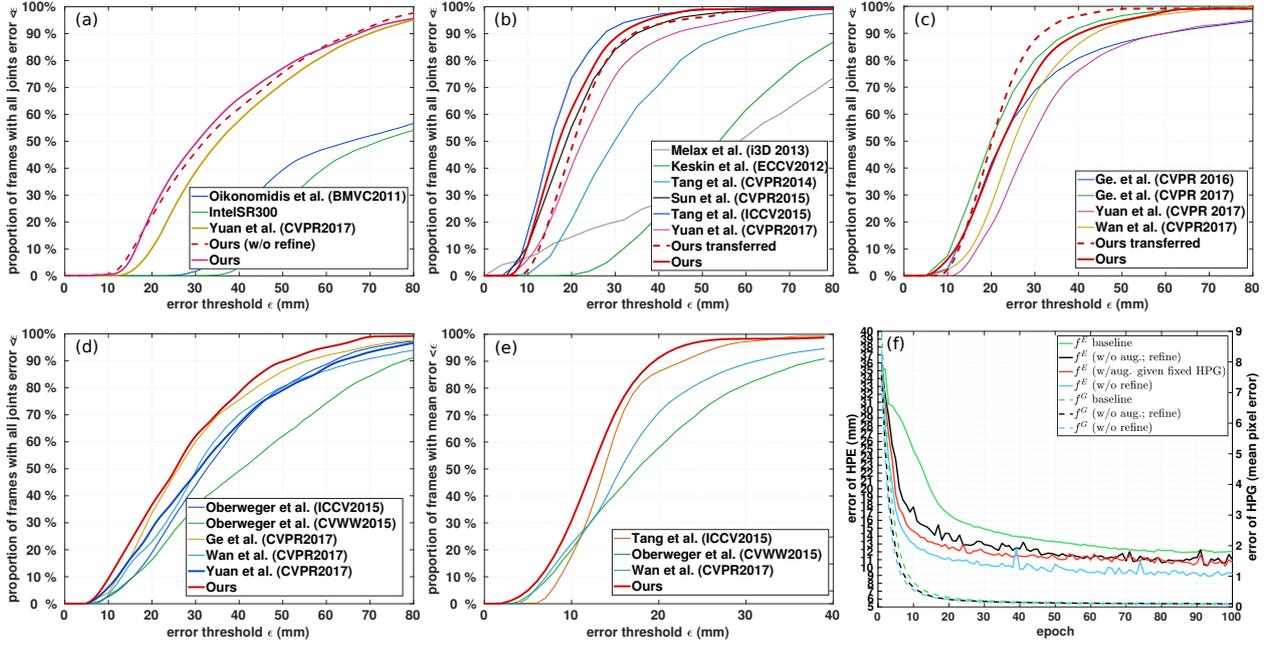

Figure 6: Accuracies of different hand pose estimators for four datasets measured in *proportion of frames with worst error<$\epsilon$* criteria (a)-(d); (e) Accuracies for *NYU*, measured in *proportion of frames with mean error<$\epsilon$* criteria (for a fair comparison with Tang *et al.* (ICCV2015)); (a-e: the larger the area under each curve is better); Ours (shape; w/o refine): our method trained with $P$, and $U$ augmented with only shape variations; Ours (rotation; w/o refine): our method trained with $P$, and $U$ augmented with only viewpoint variations; Ours (w/o refine): our method trained with $P$ and $U$ fully augmented, without refinement at testing; Ours: our final algorithm including data augmentation and the refinement step; Ours transferred: our method trained on *Big Hand 2.2M* dataset and tested on the respective dataset (see the cross-dataset experiments paragraph); (f) The 2D plot for test errors of HPE and HPG in the same epoch (trained on *ICVL*). We note strong correlations in the HPE and HPG errors.

Table 1: Evaluation of design choices: (a) Test errors of our HPG (unitless) and HPE (in mm) under varying design conditions: $f^G$ (baseline) and $f^E$ (baseline): HPG and HPE trained independently on the paired dataset $P$, respectively; $f^E$ (w/o aug.; refine): HPE trained only on $P$ pairs (Algorithm 1); $f^E$ and $f^E$ (w/o refine): HPEs trained (with skeleton augmentation) with and without the refinement step at testing, respectively; $f^G$ (w/o aug.; refine) and $f^G$ (w/o refine): HPGs trained jointly with $f^E$ (w/o aug.; refine) and $f^E$, respectively. (b) Test error of HPE on *Big Hand 2.2M* with varying numbers and types of skeleton augmentation.

| | Configuration | Big Hand 2.2M | ICVL | MSRA | NYU | | Configuration | Error (mm) |
|---|---|---|---|---|---|---|---|---|
| (a) | $f^G$ (baseline) | 0.151 | 0.588 | 0.482 | 0.451 | (b) | HPE baseline | 17.1 |
| | $f^G$ (w/o aug.; refine) | 0.124 | 0.516 | 0.470 | 0.415 | | Ours (w/o aug.; refine) | 15.7 |
| | $f^G$ (w/o refine) | 0.102 | 0.486 | 0.438 | 0.396 | | Ours (w/ in-plane-rot 10x.; w/o aug.; refine) | 14.9 |
| | $f^E$ (baseline) | 17.1 | 12.1 | 16.3 | 17.3 | | Ours (5× aug.; w/o refine) | 15.1 |
| | $f^E$ (w/o aug.; refine) | 15.7 | 10.4 | 14.4 | 16.4 | | Ours (10× aug.; w/o refine) | 14.1 |
| | $f^E$ (w/o refine) | 14.1 | 9.1 | 13.1 | 14.9 | | Ours (20× aug.; w/o refine) | 14.0 |
| | $f^E$ | 13.7 | 8.5 | 12.5 | 14.1 | | Ours (w/ in-plane-rot; 10x aug.; w/o refine) | 12.5 |

**System evaluation.** We use a commonly used criteria for hand pose estimates [47]: the *proportion (in %) of the frames with all joints error<$\epsilon$* (in Euclidean distance per joint) being smaller than a tolerance parameter $\epsilon$. Figure 6 shows the results: For all benchmarks, our algorithm ('Ours' in Fig. 6 and Table 1(a)) constantly improved upon the baseline HPE (Yuan *et al.*'s CNN estimator [59]) by a significant margin. In comparison to Tang *et al.*'s algorithm (Tang *et al.* (ICCV 2015)) [43], our algorithm shows higher and lower accuracies on *NYU* (Fig. 6(e)) and *ICVL* (Fig. 6(b)), respectively confirming the complementary nature of the two approaches. On *MSRA* containing a wide range of camera views but limited shapes and poses (see Sec. 3), Ge *et al.*'s multi-view-based approach (Ge *et al.* CVPR 2017) [11] achieved the best results, followed by our algorithm. Overall, our algorithm outperforms or is on par with state-of-the-art methods.

**Cross-dataset experiments.** In principle, the space of (augmented) skeletons is independent of specific datasets and representations and therefore, it can be shared across multiple benchmark datasets. We tested this possibility by applying our model trained on *Big Hand 2.2M* to *ICVL* and *MSRA* datasets: For *MSRA*, our model trained only on *Big Hand 2.2M* ('Ours transferred' in Fig. 6(c)) achieved the best results outperforming Ge *et al.*'s state-of-the-art model [11]: *MSRA* is limited in the range of poses (only 17 gestures), which can be compensated by transferring skeletons augmented from the much larger *Big Hand 2.2M* dataset. For *ICVL*, the transferred version is slightly worse than the original but still outperforms several existing algorithms.

**Evaluation of design choices.** Our approach enables 1) to easily augment skeleton datasets and 2) to transfer them consistently to depth maps. This approach was facilitated by training the HPE and HPG in a single unified criteria guided by the paired ($P$) and unpaired ($U$) data. To gain an insight into the contribution of each algorithm component, we evaluated the corresponding variations of our final algorithm: Table 1(a) shows that each component of our algorithm indeed makes a significant contribution to building a system as a whole: Skeletal data augmentation helps improve the performances of both HPG and HPE. Even without data augmentation, jointly training HPE and HPG within our framework, already improves the performance. Refining the prediction during testing as guided by HPG and HPD$_Y$ plus multi-view synthesis, further significantly improves the pose estimation accuracy. Figure 6(f) confirms the importance of joint HPE/HPG training: A simpler data augmentation alternative to our joint training approach is to hold the HPG $f^G$ trained on $P$ and fixed, then individually train HPE on the resulting augmented skeletons $U$ and transferred depth maps $f^G(U)$ (HPE w/aug. given fixed HPG), while this approach improves upon the HPE trained on $P$ (HPE baseline), our final algorithm shows much more significant improvements.

**Influence of the size and type of skeleton augmentation.** Table 1(b) shows that the HPE test error constantly decreases as the augmented dataset grows, confirming the importance of dataset augmentation. The accuracy gain saturates when the augmented set is around 10 times larger than the original suggesting $M = 10$ as a good trade-off between the (training) computational efficiency and accuracy. Finally, we observe that our approach of skeleton-based data augmentation and transfer is complementary to traditional view-dependent data augmentation approaches: The straightforward in-plane rotation approach applied to the skeleton and depth map pairs also significantly improve the performance of hand pose estimation, and combining the two approaches further boosts the accuracy.

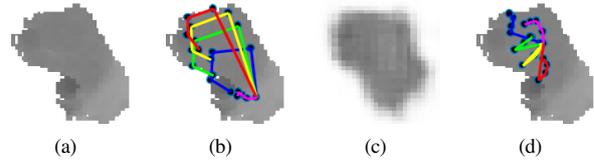

Figure 7: A failure example (*MSRA* dataset): (a) input depth map, (b) ground-truth skeleton overlaid on the input, (c) new depth map synthesized by our generator based on the ground-truth skeleton, and (d) skeleton estimated by our hand pose estimator overlaid on the depth map. When the input represents a significantly different skeletal pose from the database, the corresponding synthesized depth map (c) is blurry even when based on the ground truth, leading to a large pose estimation error (d).

## 5. Conclusions

Existing depth-based hand (pose estimation) datasets are limited in their extent in shapes, poses, and/or camera viewpoints. Traditional data augmentation approaches directly manipulate the depth map and skeleton pairs and therefore, their augmentation capabilities are limited to simple 2D view-dependent manipulations. We introduced a framework that extends this domain to a variety of hand shapes and poses. Our algorithm enables to augment data only in the skeleton space where data manipulation is intuitively controlled and greatly simplified and thereafter, automatically transfers them to realistic depth maps. This was made possible by jointly training the hand pose estimator and hand pose generator in a single unified framework. The resulting algorithm significantly outperforms or is on par with state-of-the-art hand pose estimation algorithms.

Our skeleton augmentation process enables the generator (and the corresponding pose estimator) to absorb a wide range of skeleton variations. However, when the input test entries exhibit significantly-different skeletal poses from any of the (original+augmented) training database entries, the corresponding synthesized depth maps tend to be blurry, indicating an ambiguity. This can eventually lead to pose estimation errors (Fig. 7). Future work should address this, *e.g.*, by *actively* sampling such difficult poses in the augmented skeleton space during training.

**Acknowledgments.** This work is partially supported by Huawei Technologies Co., Ltd. Kwang In Kim thanks EPSRC EP/M00533X/2 and RCUK EP/M023281/1.